\documentclass[11pt]{article}

\PassOptionsToPackage{table}{xcolor}

\usepackage[preprint]{acl}

\usepackage{times}
\usepackage{latexsym}
\usepackage[T1]{fontenc}
\usepackage[utf8]{inputenc}
\usepackage{microtype}
\usepackage{inconsolata}
\usepackage{graphicx}
\usepackage{stfloats}
\usepackage{float}
\usepackage{booktabs}
\usepackage{tabularx}
\usepackage{multirow}
\usepackage{makecell}
\usepackage{xurl}

\usepackage{amsmath,amsfonts,bm}

\def\eqref#1{equation~\ref{#1}}
\def\1{\bm{1}}

\DeclareMathAlphabet{\mathsfit}{\encodingdefault}{\sfdefault}{m}{sl}
\SetMathAlphabet{\mathsfit}{bold}{\encodingdefault}{\sfdefault}{bx}{n}

\title{SeOPD: Self-Evolving LLMs via Online Policy Distillation from Self-Generated Chain-of-Thought}

\author{Xiaoshu Chen\textsuperscript{1},
  Sihang Zhou\textsuperscript{1},
  Ke Liang\textsuperscript{1},
  Xinwang Liu\textsuperscript{1}\\
  \textsuperscript{1}National University of Defense Technology\\
  }
\begin{document}

\maketitle

\begin{abstract}
% 利用标注的答案或外部环境反馈等信息作为特权信息，self on policy model （self-opd）使得大语言模型得到了极大的提升。然而，精确的答案标注以及复杂环境的构建通常需要耗费大量的人力/计算资源。
% 因此，在本文中我们探索了如何在不依赖任何外部特权信息的情况下让模型实现自进化。 具体来说，同一个大语言模型通常会有不同的推理模型，比如深度思考以及非思考。深度思考模式的能力通常会远超非思考的。据此，我们提出了自进化的OPD（SeOPD）。SeOPD总体上分为三步，首先LLMs在深度思考rollout出思维链，然后LLMs在非思考模式下rollout出response。最后将问题、第一步的思维链+第二步的答案拼接后输入到深度思考模式的LLMs中得到深度思考模式下模型对于非思考模型输出的response中每个token的打分。通过这样的方式，SeOPD将CoT中的通用推理知识内化到模型参数中，同时提升了模型深度思考以及非思考的性能，实现了不借助任何外部信息情况下模型的自进化。在三个不同的架构的LLMs，4种不同的模型应用场景下，我们验证了SeOPD的有效性。

Recent advances in online policy self-distillation (OPSD) have demonstrated that large language models (LLMs) can improve their capabilities by leveraging external privileged information (PI), such as manual annotations or feedback from external environments. However, obtaining accurate annotations and constructing sophisticated environments often require substantial human effort and computation, limiting the scalability of OPSD. While a few recent studies have explored self-improvement without external PI, the resulting gains remain limited. In this work, we explore whether LLMs can achieve comparable self-improvement without external PI. Our key observation is that a single LLM can support multiple reasoning modes, such as deep-thinking and non-thinking modes, with deep thinking generating additional information during reasoning. Based on this observation, we propose Self-Evolving Online Policy Distillation (SeOPD), which enables LLMs to distill and internalize information generated by their own chain of thought (CoT). Specifically, it (1) generates CoT with the deep-thinking mode, (2) produces responses with the non-thinking mode, and (3) uses the generated CoT as PI to provide token-level supervision for the non-thinking response, allowing new information inferred during reasoning to guide the non-thinking mode and be internalized into the shared model parameters, thereby improving both non-thinking and deep-thinking capabilities. Extensive experiments across LLMs and tasks demonstrate the effectiveness of SeOPD.
\end{abstract}

\section{Introduction}
\begin{figure*}[t]
    \centering
    \setlength{\belowcaptionskip}{-0.4cm} %调整图片标题与下文距离
    \includegraphics[width=0.919\linewidth]{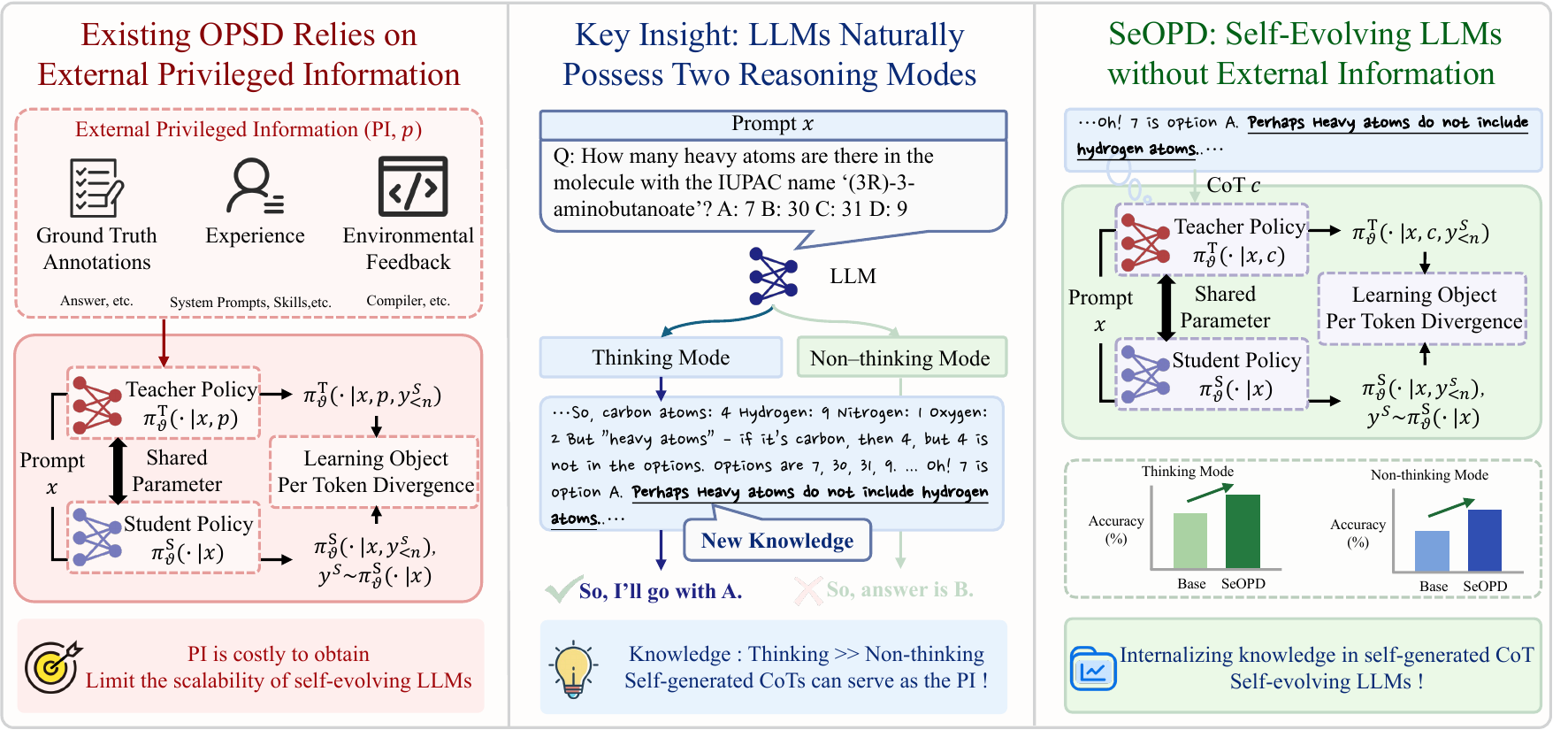}
    \caption{Motivation of SeOPD: Internalizing new knowledge from self-generated CoT for self-evolving LLMs, thereby reducing reliance on expensive external privileged information.}
    \label{fig:motivation}
\end{figure*}
% OPD的发展（OPD成为LLM后训练一种新范式。 提供token-level的奖励信号）
On-policy distillation (OPD) \citep{song2026survey} has emerged as a promising paradigm for post-training large language models (LLMs) \citep{xu2026deepseek}. It leverages a teacher policy to provide token-level supervision on trajectories sampled from the student’s own policy, thereby guiding the student toward improved behaviors. This paradigm has subsequently been extended to online policy self-distillation (OPSD) \citep{zhao2026selfdistilled}, where the teacher and student are derived from the same underlying model, offering a potentially scalable route toward LLM self-improvement.

% OPSD的做法和缺陷（内化特权信息（标注答案、经验（skill，sp，harnees）、环境反馈）的到模型参数中，但是特权信息构建昂贵
Specifically, existing OPSD approaches typically rely on privileged information (PI) that is external to the model itself. Such information provides the teacher with signals that are unavailable to the student during its rollout, allowing the teacher to distinguish desirable behaviors and generate token-level supervision. Depending on the setting, PI may take the form of manually annotated ground-truth answers ~\citep{hubotter2026reinforcement, zhao2026selfdistilled, shenfeld2026selfdistillation}, accumulated human experience (skills, system prompts, and tool-use harnesses, etc.) ~\citep{wang2026skillsd, huang2026skillconditioned, ye2026onpolicycontext}, or feedback obtained from external environments ~\citep{hubotter2026reinforcement} such as simulators and compilers. The resulting teacher can then evaluate the student trajectory conditioned on the PI and thus internalize the knowledge in PI into the model parameters. 

While effective, this paradigm introduces a fundamental scalability bottleneck: the quality of self-evolution is bounded by the availability of the PI, yet obtaining such information can be expensive. Accurate annotations require human effort, useful skills and prompts often require careful engineering, and environment-based feedback may require sophisticated environments, tools or substantial computation.

While recent efforts \citep{li2026policy} have explored self-evolution without relying on external PI, the resulting improvements remain limited, suggesting that simply removing external supervision does not necessarily provide a reliable source of useful information for self-evolution. We therefore address this challenge by looking inside the LLM itself. Modern LLMs increasingly support multiple reasoning modes ~\citep{qwen3, tencent2026hy4, anthropic2026claudeopus48}\, such as deep-thinking and non-thinking modes, within the same model. Importantly, these modes are not merely different ways of formatting a response: they can provide different amounts of information during inference. As illustrated in the middle part of Figure \ref{fig:motivation}, given the same problem, the deep-thinking mode can generate an explicit chain-of-thought (CoT) through which the model infers new information or knowledge by comparing intermediate reasoning results with the candidate options, concluding that hydrogen does not belong to the category of heavy atoms. This observation points to an overlooked source of PI: the model's own reasoning process. By using the information generated during deep thinking to guide the non-thinking mode, the model can internalize such newly inferred information into its parameters, providing a pathway for self-evolution without external PI.

Based on this insight, we propose Self-Evolving Online Policy Distillation (SeOPD), a framework that enables an LLM to distill its own reasoning capabilities without any external privileged information. The key idea is to use the model's thinking mode as a teacher and its non-thinking mode as the student. Specifically, SeOPD first samples a CoT trajectory from the thinking mode, providing a self-generated CoT for the given question. The same model then independently produces a response in the non-thinking mode. Finally, we feed the question, the self-generated CoT, and the non-thinking response back into the thinking-mode policy. Conditioned on the CoT, the thinking teacher can provide token-level evaluations of the non-thinking response, yielding an online distillation signal without requiring external PI. 

An important consequence of SeOPD is that the optimization is not restricted to improving the non-thinking policy. By distilling reasoning-conditioned knowledge into the shared parameters of the model, the model can simultaneously improve its non-thinking behavior and its underlying reasoning capability. In other words, SeOPD turns the model's heterogeneous modes into mutually beneficial learning signals: thinking mode provides supervision for non-thinking mode, while the resulting parameter update further strengthens the thinking mode itself. 
%This establishes a new form of self-evolution in which the source of privileged information is no longer external to the model, but emerges naturally from the model's own reasoning process.

Our contributions are summarized as follows:

1) Self-generated CoT as privileged information. We identify self-generated CoT as a novel source of privileged supervision, enabling an LLM to leverage its own reasoning process to supervise itself.

2) Stable self-evolving training framework. We introduce SeOPD, a new OPSD framework that enables the model to steadily improve both its fast non-thinking and its deep-thinking capability. 

3) Extensive empirical validation. We verify the effectiveness of SeOPD across different LLM architectures and application scenarios.

% 贡献点总结

\section{Related Work}

\subsection{Large Language Reasoning Models}
Large language reasoning models such as OpenAI o1~\citep{openai2024o1} and DeepSeek-R1~\citep{deepseekr1} have established long chain-of-thought (CoT) reasoning as a central paradigm for complex tasks, trading inference-time computation for higher accuracy~\citep{wei2022cot,snell2025scaling,chen2026imgcot,chen2026puttingthinkinghatssurvey}. Such behavior is typically elicited via reinforcement learning with verifiable rewards (RLVR) using policy-gradient algorithms such as PPO~\citep{schulman2017ppo} and GRPO~\citep{shao2024deepseekmath}. A notable recent trend unifies thinking and non-thinking behaviors within a single model: rather than maintaining separate checkpoints (e.g., GPT-4o~\citep{openai2024gpt4o} vs.\ o1, or Qwen2.5~\citep{qwen2.5} vs.\ QwQ~\citep{qwq32b}), models such as Qwen3~\citep{qwen3} and Claude~\citep{anthropic2025claude37} expose both a \emph{deep-thinking} mode and a \emph{non-thinking} mode. Although these modes share parameters, the deep-thinking mode attains substantially higher accuracy on complex tasks~\citep{qwen3, anthropic2025claude37}. While prior work treats these modes as complementary interfaces for controlling cost, we instead exploit the \emph{information gap} between them as an internal PI for supervision.

\subsection{On-Policy Self-Distillation}
Knowledge distillation~\citep{hinton2015distilling} transfers knowledge from a stronger teacher to a weaker student~\citep{chen2025adaptive,chen-etal-2025-skip}. A key limitation of distilling on a fixed, teacher-generated corpus is the train--inference distribution mismatch. On-policy distillation (OPD)~\citep{agarwal2024onpolicy,gu2024minillm} addresses this by sampling trajectories from the student's own policy and having the teacher provide fine-grained, token-level supervision on them, making it more sample-efficient and stable than sequence-level distillation or purely reward-based RL. A natural extension is online policy self-distillation (OPSD)~\citep{hubotter2026reinforcement, zhao2026selfdistilled}, where teacher and student are instantiated from the same model, offering a scalable route to self-improvement. The central challenge is endowing the teacher with an advantage over the student, which existing methods resolve by conditioning the teacher on PI unavailable to the student during rollout: annotated ground-truth answers or hints~\citep{hubotter2026reinforcement, zhao2026selfdistilled, shenfeld2026selfdistillation, penaloza2026privileged, ding2026hdpo, zhang2026policy}, accumulated human experience such as skills~\citep{wang2026skillsd, huang2026skillconditioned}, system prompts~\citep{ye2026onpolicycontext} and harness~\citep{zhao2026training}, or feedback from external environments such as compilers and simulators~\citep{hubotter2026reinforcement}. However, these approaches share a fundamental bottleneck: the quality of self-evolution depends on external PI that is costly to obtain. Recent work on U-OPSD~\citep{li2026policy} removes this dependence by constructing a pseudo-solution from self-consistency across multiple rollouts and using it as privileged teacher context. Our work similarly derives PI internally, but instead uses the self-generated CoT from the model's own deep-thinking mode, enabling the model to supervise itself without external supervision or multiple rollouts.

\section{Method}
\label{sec:method}

\subsection{Notations}
Let $\mathcal{D}={x}$ denote a collection of input prompts, and let $\pi_{\theta}$ denote the language model policy parameterized by $\theta$. LLMs often support multiple reasoning modes. We consider two modes of the same model: a \emph{thinking mode}, which explicitly generates a CoT, and a \emph{non-thinking mode}, which directly produces the response. We denote the corresponding policies by $\pi_{\theta}^{\mathrm{T}}$ and $\pi_{\theta}^{\mathrm{S}}$ respectively.

\begin{figure*}[t]
    \centering
    \setlength{\belowcaptionskip}{-0.45cm} %调整图片标题与下文距离
    \includegraphics[width=0.7\linewidth]{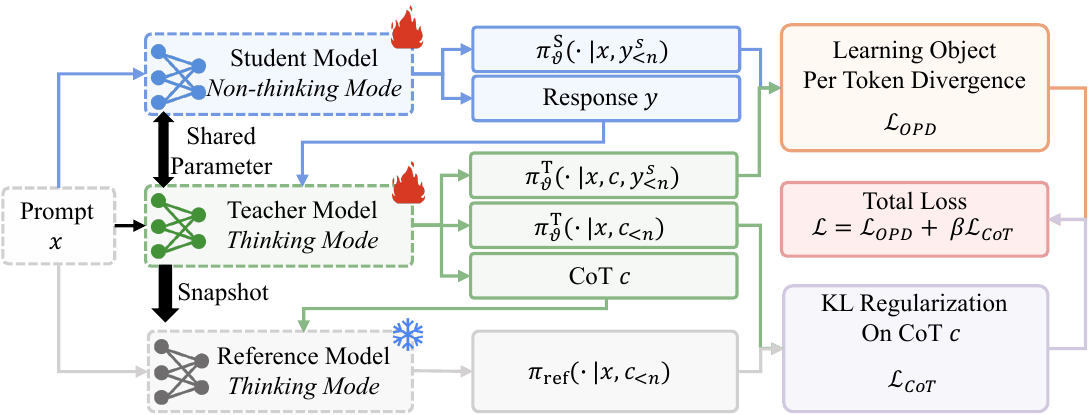}
    \caption{Overview of the SeOPD training framework}
    \label{fig:method}
\end{figure*}

\subsection{Self-Evolving Online Policy Distillation}
The overall architecture of SeOPD is illustrated in Figure~\ref{fig:method}. SeOPD consists of three key stages: \emph{teacher and student rollouts}, where the same LLM generates a CoT and a response under different reasoning modes; \emph{teacher scoring}, where the $\pi_{\theta}^{\mathrm{T}}$ uses the self-generated CoT to provide token-level supervision for the $\pi_{\theta}^{\mathrm{S}}$ response; and \emph{CoT regularization}, which constrains the evolution of the $\pi_{\theta}^{\mathrm{T}}$ with respect to a frozen $\pi_{\mathrm{ref}}$. We describe each component in detail below.
\subsubsection{Teacher and Student Rollouts}
For each training prompt $x\in\mathcal{D}$, we perform two online rollouts using the same model under different reasoning modes.

\paragraph{Student Rollout.}
Given an input $x$, the non-thinking policy $\pi_{\theta}^{\mathrm{S}}$ directly generates a response without explicitly producing a reasoning trajectory:
\begin{equation}
y=(y_1,\ldots,y_n)
\sim
\pi_{\theta}^{\mathrm{S}}(\cdot|x).
\end{equation}
For each generated prefix $y_{<n}$, the student policy induces a token-level conditional distribution
\begin{equation}
q =
\pi_{\theta}^{\mathrm{S}}
(\cdot|x,y_{<n}).
\end{equation}

\paragraph{Teacher Rollout.}
Independently, the thinking policy $\pi_{\theta}^{\mathrm{T}}$ explicitly reasons about the input and generates a CoT trajectory:
\begin{equation}
c=(c_1,\ldots,c_m)
\sim
\pi_{\theta}^{\mathrm{T}}(\cdot|x).
\end{equation}
The generated CoT $c$ captures the model's internal reasoning process and serves as self-generated PI for evaluating the student response.

\subsubsection{Teacher Scoring}

After obtaining the student response $y$ and the self-generated CoT $c$, we use the CoT to provide additional reasoning context for evaluating the student-generated response. Specifically, for each response prefix $y_{<n}$, we feed $x$, $c$, and $y_{<n}$ into the thinking policy $\pi_{\theta}^{\mathrm{T}}$, yielding the CoT-conditioned token-level distribution
\begin{equation}
p =
\pi_{\theta}^{\mathrm{T}}
(\cdot|x,c,y_{<n}).
\end{equation}

The CoT $c$ provides the thinking policy with additional reasoning context that is unavailable to the non-thinking policy. As a result, $p$ incorporates information extracted from the model's reasoning process when predicting the next response token. We use this distribution to provide token-level supervision for the student policy.

Specifically, we minimize the symmetric KL divergence between the student distribution and the CoT-conditioned teacher distribution:
\begin{equation}
\mathcal{L}_{\mathrm{OPD}}
=
\frac{1}{2}
D_{\mathrm{KL}}(q|p)
+
\frac{1}{2}
D_{\mathrm{KL}}(p|q),
\end{equation}
where $D_{\mathrm{KL}}$ refers to the KL divergence.
The resulting objective encourages the student to align its token-level predictions with the CoT-informed distribution produced by the teacher.

\subsubsection{CoT Regularization}

The thinking and non-thinking policies share the same model parameters. Consequently, optimizing the distillation objective also changes the thinking policy. To prevent the reasoning behavior from drifting excessively and thereby harming training stability, we maintain a frozen reference policy $\pi_{\mathrm{ref}}$, initialized from the model before training.

Then, we regularize the thinking policy on its self-generated CoT by constraining its token-level distribution to remain close to that of $\pi_{\mathrm{ref}}$:
\begin{equation}
\mathcal{L}_{\mathrm{CoT}}
=
\frac{1}{m}
\sum_{t=1}^{m}
D_{\mathrm{KL}}
\left(
\pi_{\theta}^{\mathrm{T}}
(\cdot|x,c_{<t})
\middle|
\pi_{\mathrm{ref}}
(\cdot|x,c_{<t})
\right).
\end{equation}

\subsubsection{Training Objective}
The final training objective is
\begin{equation}
\mathcal{L}
=
\mathcal{L}_{\mathrm{OPD}}
+
\beta \times\mathcal{L}_{\mathrm{CoT}},
\end{equation}
where $\beta$ controls the strength of the CoT regularization.

During optimization, both the CoT and the student response are dynamically sampled online. The CoT-conditioned teacher distribution enables the non-thinking policy to internalize useful information from the stronger reasoning process, improving its non-thinking capability. Meanwhile, the shared model parameters absorb recurring reasoning patterns, effective problem-solving strategies, and newly acquired knowledge emerging from CoT, thereby strengthening the underlying capabilities of the LLM itself. As the model becomes stronger, the thinking policy can generate increasingly informative CoTs, which in turn provide stronger supervision for subsequent self-evolution. The CoT regularization further stabilizes this self-evolution process by preventing excessive deviation from the initial reasoning behavior.

\section{Experiments}
In this section, we investigate four key questions: 

    \textbf{Q1} whether SeOPD can simultaneously improve thinking and non-thinking capabilities?

    \textbf{Q2} how SeOPD compares with OPSD methods using other forms of privileged information? 

    \textbf{Q3} Does SeOPD really internalize new knowledge from its self-generated CoT? 

    \textbf{Q4} whether SeOPD can benefit from CoT generated by a stronger LLM as privileged information. 

Unless otherwise specified, we report \textbf{mean@16}: for each case we rollout 16 completions and average their binary correctness. Detailed experimental settings are provided in Appendix~\ref{sec:expsetting}.

\subsection{The effectiveness of SeOPD (for Q1)}
In this section, we first evaluate the effectiveness of SeOPD for self-evolution, and then conduct a series of ablation studies to further investigate its properties.
\subsubsection{Main results}
\begin{table*}[t]
\centering
\caption{Performance comparison across science Q\&A, tool use, and coding tasks.
$\uparrow$ indicates an improvement over the corresponding base model.}
\label{tab:main_results}
\setlength{\tabcolsep}{1.8pt}
\renewcommand{\arraystretch}{1.05}

\begin{tabular}{llccccccc}
\toprule
\multirow{2}{*}{\makecell{Model}} & \multirow{2}{*}{\makecell{Mode}} & \multicolumn{5}{c}{Science Q\&A }
& \multicolumn{1}{c}{Tool Use}
& \multicolumn{1}{c}{Coding} \\
\cmidrule(lr){3-7}
\cmidrule(lr){8-8}
\cmidrule(lr){9-9}
&
& Chemistry & Physics & Biology & Materials & Mean
& ToolAlpaca & LCBv6 \\
\midrule

\multirow{2}{*}{Qwen3-4B}
& Thinking
& 45.9 & 56.2 & 29.0 & 73.1 & 51.1 & 57.3 & 41.7 \\
& Non-Thinking
& 42.9 & 60.6 & 33.9 & 62.6 & 50.0 & 58.4 & 46.6 \\

\addlinespace[1pt]

\multirow{2}{*}{\makecell{+ SeOPD}}
& \cellcolor{gray!15} Thinking
& \cellcolor{gray!15}\textbf{59.3}$\uparrow$
& \cellcolor{gray!15}\textbf{68.1}$\uparrow$
& \cellcolor{gray!15}\textbf{39.9}$\uparrow$
& \cellcolor{gray!15}72.8
& \cellcolor{gray!15}\textbf{60.0}$\uparrow$
& \cellcolor{gray!15}\textbf{59.5}$\uparrow$
& \cellcolor{gray!15} \textbf{63.5}$\uparrow$ \\
& \cellcolor{gray!15} Non-Thinking
& \cellcolor{gray!15}\textbf{58.7}$\uparrow$
& \cellcolor{gray!15}\textbf{61.7}$\uparrow$
& \cellcolor{gray!15}\textbf{40.1}$\uparrow$
& \cellcolor{gray!15}\textbf{75.7}$\uparrow$
& \cellcolor{gray!15}\textbf{59.7}$\uparrow$
& \cellcolor{gray!15}\textbf{62.0}$\uparrow$
& \cellcolor{gray!15} \textbf{53.4}$\uparrow$ \\

\midrule

\multirow{2}{*}{Qwen3-8B}
& Thinking
& 47.5 & 58.0 & 26.9 & 62.8 & 48.8 & 58.4 & 43.1 \\
& Non-Thinking
& 41.2 & 60.0 & 28.6 & 60.0 & 47.5 & 58.7 & 51.7 \\

\addlinespace[1pt]

\multirow{2}{*}{\makecell{+ SeOPD}}
& \cellcolor{gray!15} Thinking
& \cellcolor{gray!15}\textbf{63.5}$\uparrow$
& \cellcolor{gray!15}\textbf{66.5}$\uparrow$
& \cellcolor{gray!15}\textbf{35.1}$\uparrow$
& \cellcolor{gray!15}\textbf{75.5}$\uparrow$
& \cellcolor{gray!15}\textbf{60.2}$\uparrow$
& \cellcolor{gray!15}\textbf{61.3}$\uparrow$
& \cellcolor{gray!15}\textbf{64.5}$\uparrow$ \\
& \cellcolor{gray!15} Non-Thinking
& \cellcolor{gray!15}\textbf{65.8}$\uparrow$
& \cellcolor{gray!15}\textbf{65.5}$\uparrow$
& \cellcolor{gray!15}\textbf{36.6}$\uparrow$
& \cellcolor{gray!15}\textbf{74.6}$\uparrow$
& \cellcolor{gray!15}\textbf{60.6}$\uparrow$
& \cellcolor{gray!15}\textbf{63.1}$\uparrow$
& \cellcolor{gray!15}\textbf{56.4}$\uparrow$ \\

\midrule

\multirow{2}{*}{
\begin{tabular}[l]{@{}l@{}}
DeepSeek-Distill\\
Qwen3-8B
\end{tabular}}
& Thinking
& 46.3 & 61.7 & 22.1 & 62.9 & 48.3 & 38.1 & 20.8 \\
& Non-Thinking
& 46.8 & 61.8 & 32.6 & 66.6 & 51.9 & 47.9 & 44.9 \\

\addlinespace[1pt]

\multirow{2}{*}{\makecell{+ SeOPD}}
& \cellcolor{gray!15} Thinking
& \cellcolor{gray!15}\textbf{59.5}$\uparrow$
& \cellcolor{gray!15}\textbf{68.2}$\uparrow$
& \cellcolor{gray!15}\textbf{26.8}$\uparrow$
& \cellcolor{gray!15}\textbf{67.2}$\uparrow$
& \cellcolor{gray!15}\textbf{55.4}$\uparrow$
& \cellcolor{gray!15}\textbf{52.9}$\uparrow$
& \cellcolor{gray!15}\textbf{49.6}$\uparrow$ \\
& \cellcolor{gray!15} Non-Thinking
& \cellcolor{gray!15}\textbf{63.6}$\uparrow$
& \cellcolor{gray!15}\textbf{66.1}$\uparrow$
& \cellcolor{gray!15}28.8
& \cellcolor{gray!15}\textbf{73.6}$\uparrow$
& \cellcolor{gray!15}\textbf{58.0}$\uparrow$
& \cellcolor{gray!15}\textbf{54.7}$\uparrow$
& \cellcolor{gray!15}\textbf{50.0}$\uparrow$ \\

\bottomrule
\end{tabular}
\end{table*}

Table~\ref{tab:main_results} summarizes SeOPD across science Q$\&$A, tool use, and coding tasks in both Thinking and Non-Thinking modes. SeOPD consistently improves both modes across diverse tasks and model families. The gains in Non-Thinking demonstrate the effectiveness of self-generated CoT as privileged information, while the improvement in Thinking suggests that the model can internalize useful knowledge from its own CoT. Overall, SeOPD enables self-evolution across tasks and reasoning modes without external privileged information.

\begin{figure*}[t]
\centering
\setlength{\belowcaptionskip}{-0.25cm} %调整图片标题与下文距离
\begin{minipage}[t]{0.60\textwidth}
  \centering
  \includegraphics[width=\linewidth]{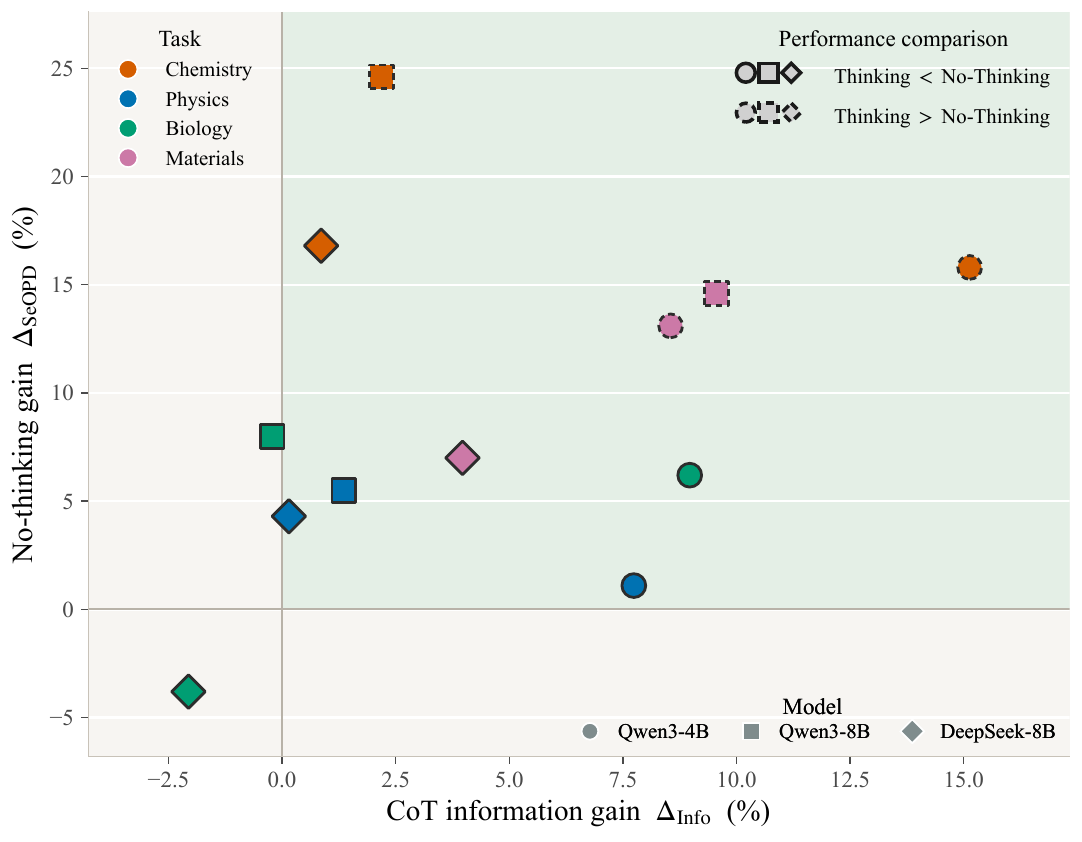}
  \caption{CoT information gain versus Non-Thinking improvement.
Positive gain improves every task. Negative gain means Thinking fails to guide Non-Thinking, except Qwen3-8B on Biology, where the mean is slightly negative but most cases stay positive.}
  \label{fig:info_gain_correlation}
\end{minipage}
\hfill
\begin{minipage}[t]{0.38\textwidth}
  \centering
  \includegraphics[width=\linewidth]{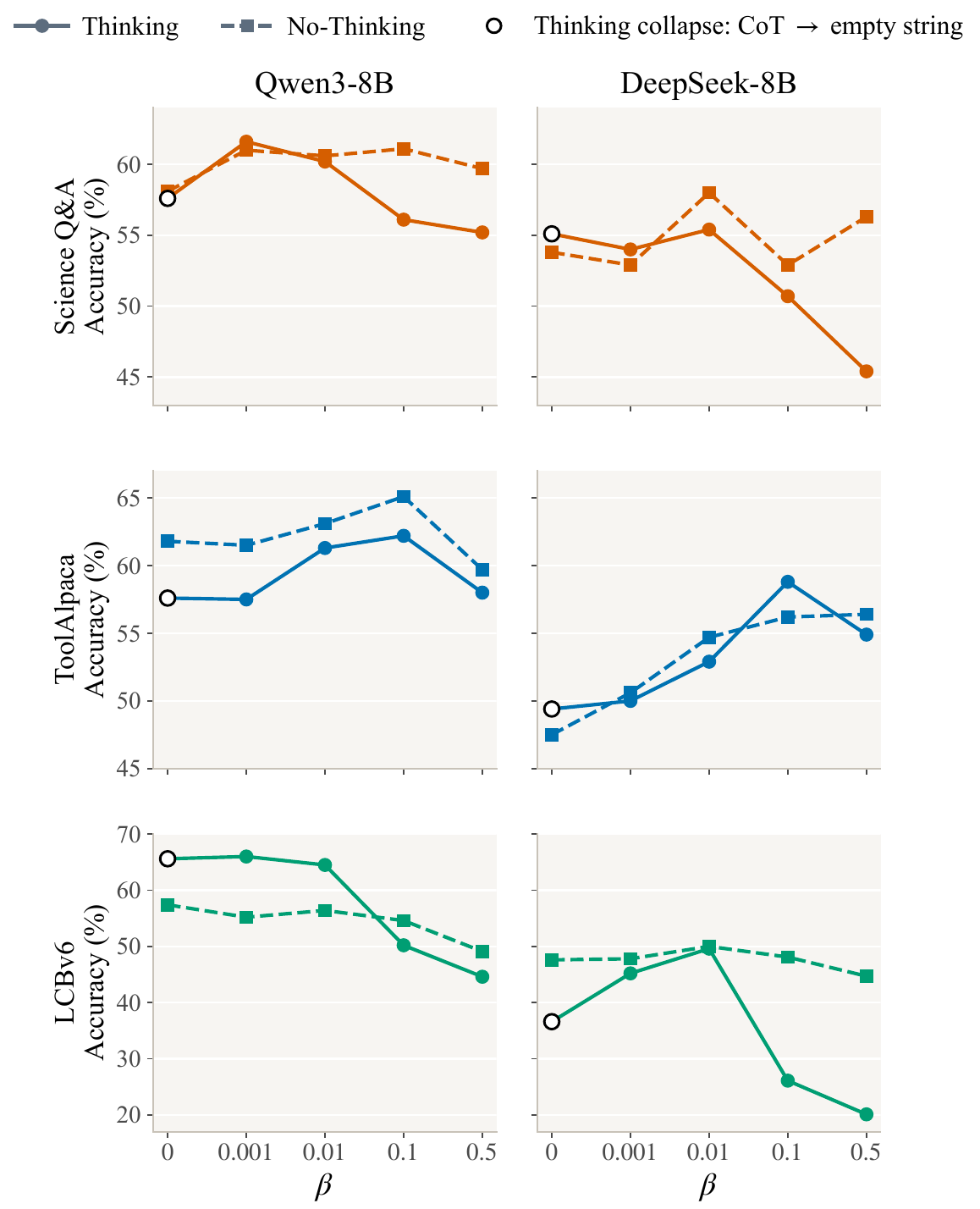}
    \caption{Effect of $\beta$ on Self-Evolution. An intermediate regularization strength balances informative CoT and knowledge internalization.}
    \label{fig:beta}
\end{minipage}
\end{figure*}

% \subsubsection{Why Bad Teachers Can Successfully Guide Good Students}
% \begin{figure}[ht]
%     \centering
%     \includegraphics[width=\linewidth]{imgs/info_gain_scatter.pdf}
%     \caption{CoT information gain versus No-Thinking improvement after SeOPD.
% Color indicates the task and shape indicates the model.
% Solid and dashed borders mark whether the initial Thinking policy is weaker
% or stronger than No-Thinking, respectively.
% The overall fit shows a moderate positive correlation ($r=0.34$).
% Positive information gain consistently yields positive SeOPD gains, whereas
% non-positive information gain leads to mixed and unstable evolution.}
%     \label{fig:info_gain_correlation}
% \end{figure}

\subsubsection{Why Bad Teachers Can Successfully Guide Good Students}
A natural question is why SeOPD can improve the Non-Thinking policy even when
the initial Thinking policy is weaker in accuracy in Table~\ref{tab:main_results}.
We hypothesize that teacher accuracy and informativeness are not equivalent.
Thus, we measure the latter by the CoT information gain:
$\Delta_{\mathrm{Info}} = P(y\mid x,c) - P(y\mid x)$,
where $x$, $c$, and $y$ denote the input, generated CoT, and correct answer.
We report this gain on science Q\&A, since its multiple-choice format places the answer in a single token and makes the gain easy to compute.

Figure~\ref{fig:info_gain_correlation} shows that every setting with $\Delta_{\mathrm{Info}}>0$ improves Non-Thinking, while a negative gain leaves Thinking unable to guide Non-Thinking, as on DeepSeek-8B Biology, where its accuracy drops. Qwen3-8B on Biology looks like a counterexample, since its mean gain is slightly negative yet Non-Thinking still improves. However, the gain is positive on $65.7\%$ of its trainging samples, and only a few strongly negative cases drag the mean below zero, so the gain remains positive on most cases and the improvement is consistent with our observation.

These results confirm our hypothesis that the new information in the CoT, rather than the accuracy of Thinking, is what drives the improvement of Non-Thinking. Consequently, as long as its CoT provides useful information, SeOPD can still drive self-evolution.

\subsubsection{Holdout tasks}
To examine catastrophic forgetting, we train the model exclusively on LCBv6
and evaluate it on the unseen IFEval \citep{ifeval2023} and MATH-500 \cite{mathdataset2021} tasks.
As shown in Table~\ref{tab:holdout}, SeOPD substantially improves both
Thinking and Non-Thinking performance on LCBv6, while causing only minor changes
on the holdout tasks. These results suggest that SeOPD can acquire new task-specific capabilities while largely preserving existing ones.
% To examine whether SeOPD causes catastrophic forgetting, we train the model exclusively on LCBv6 and evaluate it on the unseen IFEval and MATH-500 tasks. As shown in Table~\ref{tab:holdout}, SeOPD substantially improves both Thinking and No-Thinking performance on LCBv6, while causing only minor changes on the two holdout tasks. Notably, the No-Thinking performance on MATH-500 even improves slightly.

% These results indicate that SeOPD can acquire new task-specific capabilities while largely preserving the model's existing capabilities, without causing significant catastrophic forgetting.

\begin{table*}[t]
\centering
\caption{Performance after training only on LCBv6. IFEval and MATH-500 are hold-out tasks.}
\label{tab:holdout}
\setlength{\tabcolsep}{3.2pt}
\renewcommand{\arraystretch}{1.05}
\begin{tabular}{lcccccc}
\toprule
\multirow{2}{*}{Method}
  & \multicolumn{2}{c}{LCBv6}
  & \multicolumn{2}{c}{IFEval}
  & \multicolumn{2}{c}{MATH-500} \\
\cmidrule(lr){2-3}
\cmidrule(lr){4-5}
\cmidrule(lr){6-7}
  & Thinking & Non-Thinking
  & Thinking & Non-Thinking
  & Thinking & Non-Thinking \\
\midrule
Qwen3-8B
  & 43.1 & 51.7
  & \textbf{85.0} & \textbf{83.1}
  & \textbf{97.1} & 83.8 \\

\rowcolor{gray!15}
SeOPD (LCBv6)
  & \textbf{64.5} & \textbf{56.4}
  & 84.6 & 81.5
  & 96.4 & \textbf{84.2} \\
\bottomrule
\end{tabular}
\end{table*}

\subsubsection{The influence of the CoT regularization term}

% \begin{figure*}[ht]
%     \centering
%     \includegraphics[width=\linewidth]{imgs/beta_sensitivity.pdf}
%     \caption{Effect of $\beta$ on Self-Evolution}
%     \label{fig:beta}
% \end{figure*}

We investigate the effect of the CoT regularization weight $\beta$ by varying it over ${0, 0.001, 0.01, 0.1, 0.5}$. As shown in Figure~\ref{fig:beta}, the Thinking policy exhibits a consistent non-monotonic trend, generally improving and then deteriorating as $\beta$ increases.

\begin{figure}[t!]
    \centering
    \setlength{\belowcaptionskip}{-0.3cm} %调整图片标题与下文距离
    \includegraphics[width=0.75\linewidth]{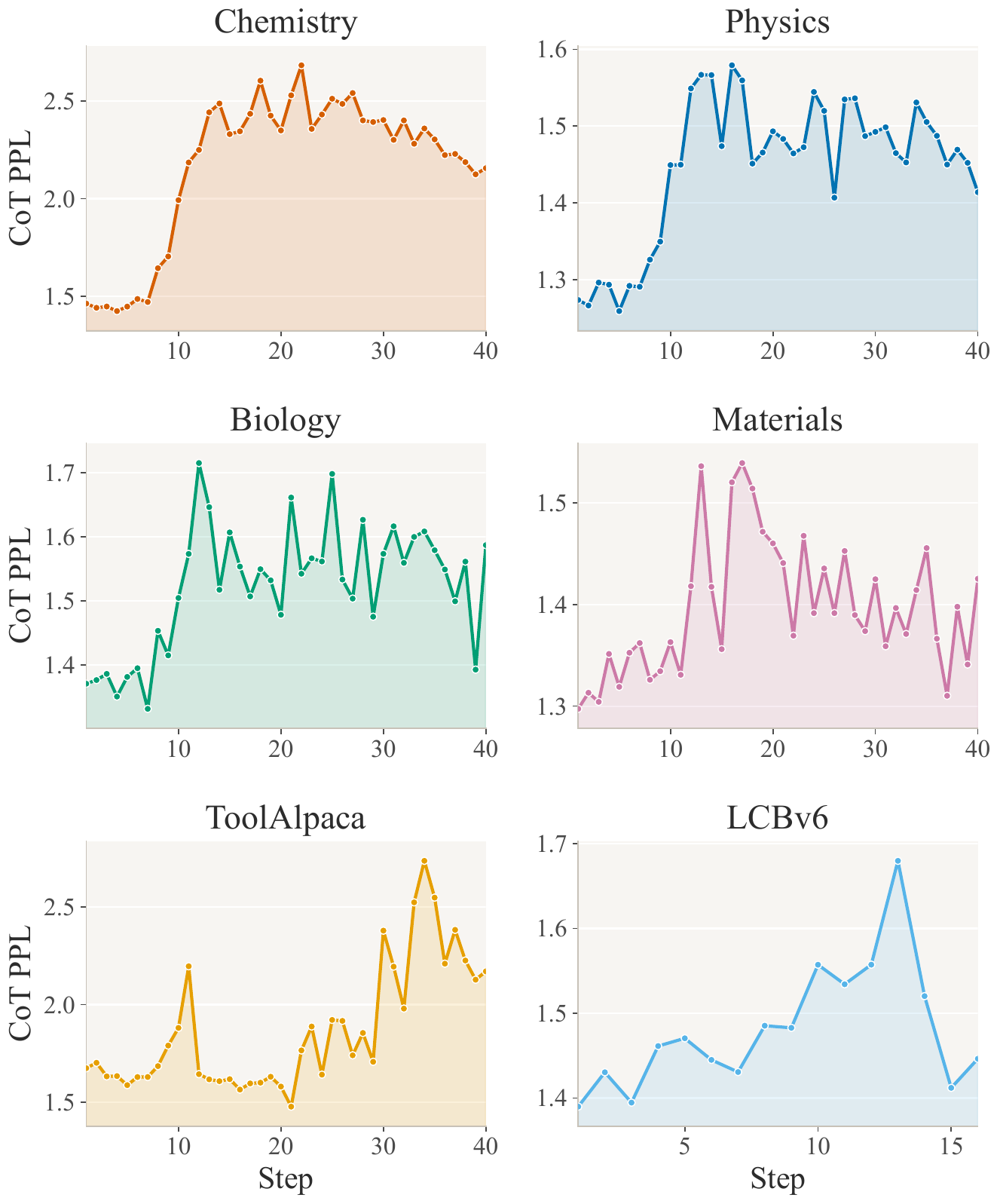}
    \caption{Per-step CoT perplexity under the frozen reference model.
Rising PPL across domains indicates that SeOPD internalizes new knowledge
and, during reasoning, uses it to produce additional knowledge that the
initial model cannot understand.}
    \label{fig:ppl_refkl}
\end{figure}

When $\beta=0$, the Thinking policy gradually stops generating explicit CoT and directly produces the final response, causing the reasoning trajectory used as privileged information for the Non-Thinking policy to disappear. A moderate $\beta$ alleviates this degeneration by preserving informative reasoning trajectories, which benefits both the Thinking policy and subsequent distillation. However, excessively large $\beta$ overly constrains the policy to existing reasoning behavior, thus hindering self-evolution. The Non-Thinking policy shows a similar trend, further suggesting that CoT regularization should balance reasoning preservation with policy evolution.

Overall, these results reveal a trade-off between \emph{reasoning preservation} and \emph{policy evolution}: insufficient regularization causes CoT collapse, while excessive regularization restricts the model's ability to internalize new information. An intermediate $\beta$ provides a better balance between maintaining informative CoT trajectories and enabling self-evolution.

\subsection{Comparison of different PIs (for Q2)}

\begin{table*}[t]
\centering
\caption{Comparison of external and internal PI on Science Q\&A, tool use, and coding. All experiments are conducted on Qwen3-8B. ``--'' indicates that the method is not applicable to the corresponding setting.}
\label{tab:baseline_results}
\setlength{\tabcolsep}{1.8pt}
\renewcommand{\arraystretch}{1.05}
\begin{tabular}{lccccccc}
\toprule
\multirow{2}{*}{Privileged Information} & \multicolumn{5}{c}{Science Q\&A} & Tool Use & Coding \\
\cmidrule(lr){2-6} \cmidrule(lr){7-7} \cmidrule(lr){8-8}
 & Chemistry & Physics & Biology & Materials & Mean & ToolAlpaca & LCBv6 \\
\midrule
\multicolumn{8}{l}{\textit{External privileged information}} \\
Answer (SDPO)
  & \textbf{80.1} & \textbf{77.3} & \textbf{60.0} & \textbf{78.0} & \textbf{73.9} & -- & -- \\
Environment feedback (SDPO)
  & -- & -- & -- & -- & -- & 64.71 & \textbf{62.7} \\
Few-shot (SDFT)
  & 75.9 & 73.2 & 53.5 & 76.5 & 69.8 & \textbf{65.07} & 53.4 \\
System Prompt (OPCD)
  & 47.0 & 57.4 & 37.8 & 68.2 & 52.6 & 57.0 & 52.5 \\
Skills
  & -- & -- & -- & -- & -- & -- & 51.3 \\
\midrule
\multicolumn{8}{l}{\textit{Internal privileged information}} \\
Self-consistency (U-OPSD)
  & 49.3 & 63.6 & 30.1 & 57.0 & 50.0 & 63.5 & 52.9 \\
\rowcolor{gray!15}
Self-generated CoT (\textbf{SeOPD})
  & 65.8 & 65.5 & 36.6 & 74.6 & 62.4 & 63.1 & 61.4 \\
\bottomrule
\end{tabular}
\end{table*}

To investigate the effect of different sources of privileged information (PI), we compare methods using external and internal PI. External PI includes ground-truth answers or environment feedback (SDPO) \citep{hubotter2026reinforcement}, few-shot demonstrations continuously refined by an external language model (SDFT) \citep{shenfeld2026selfdistillation}, tuned system prompts (OPCD) \citep{ye2026onpolicycontext}, and task-specific skills \citep{miaoge2026competitiveprogrammingexpert}. Internal PI is derived solely from the model itself: U-OPSD \citep{li2026policy} uses response consistency, while SeOPD uses self-generated CoT. Detailed descriptions of all methods are provided in the Appendix \ref{sec:other_method}.

As shown in Table~\ref{tab:baseline_results}, methods using external PI generally achieve stronger absolute performance, particularly when direct correctness signals or task-specific knowledge are available. Nevertheless, SeOPD achieves competitive performance with some external-PI methods, outperforming both the task-specific skill and tuned system-prompt approaches while relying solely on internal PI.

SeOPD also substantially outperforms U-OPSD, which relies solely on internal PI based on response-level consistency. This suggests that reasoning-level information provides more informative supervision than response-level consistency. Overall, these results demonstrate that self-generated CoT can serve as an effective source of internal PI for self-evolution.

\subsection{Internalize knowledge within the CoT (for Q3)}

\begin{figure*}[t]
    \centering
    \setlength{\belowcaptionskip}{-0.3cm} %调整图片标题与下文距离
    \includegraphics[width=0.75\linewidth]{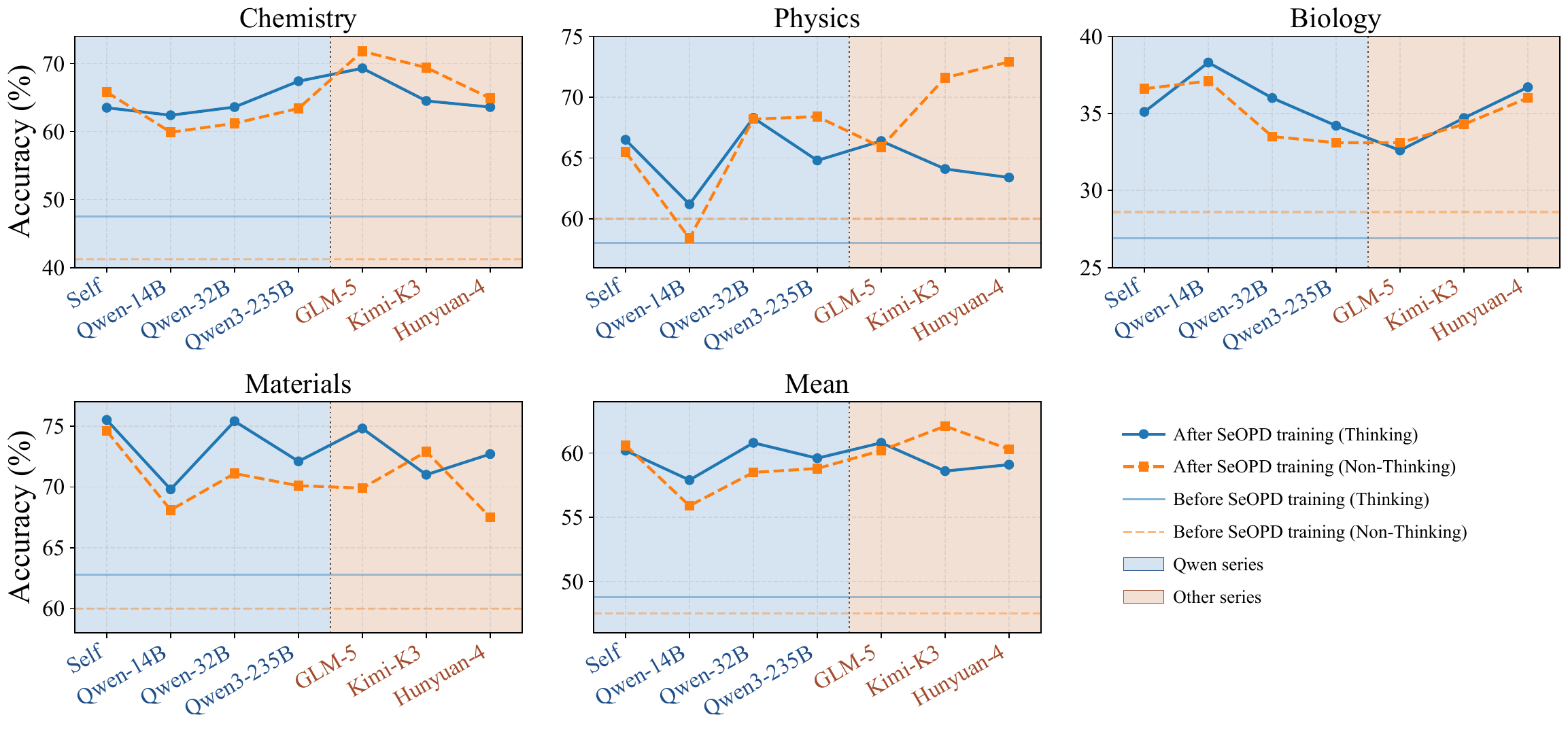}
    \caption{Comparison of different CoT for SeOPD on Science Q$\&$A. Experiments are conducted on Qwen3-8B.}
    \label{fig:cot_sources}
\end{figure*}

\begin{table}[t]
\centering
\caption{Comparison between the reference model and SeOPD using the same CoT from the reference model.}
\label{tab:ref_cot}
\renewcommand{\arraystretch}{1.18}
\small
\begin{tabularx}{\linewidth}{@{}>{\raggedright\arraybackslash}Xccc@{}}
\toprule
Method & \makecell{Science\\Q\&A} & \makecell{Tool\\Use} & Coding \\
\midrule
Reference Model (Qwen3-8B)
  & 48.8 & 58.4 & 43.1 \\
\rowcolor{gray!15}
After SeOPD
  & \textbf{52.4} & \textbf{58.9} & \textbf{45.8} \\
\bottomrule
\end{tabularx}
\end{table}

To investigate whether SeOPD can acquire and internalize new information from its self-generated CoT, we examine both its performance gains and the evolution of its generated reasoning traces.

First, we fix the CoT generated by the Reference Model, so that both models are conditioned on exactly the same CoT. As shown in Table~\ref{tab:ref_cot}, SeOPD consistently outperforms the Reference Model across Science Q$\&$A, Tool Use, and Coding. Since the two models use the same CoT, these gains cannot be attributed to better reasoning trajectories. Instead, they suggest that SeOPD can learn and internalize useful information from the CoT, resulting in stronger underlying model capabilities.

We further examine whether SeOPD progressively generates information beyond the representational scope of the initial model. At different training stages, we collect CoTs generated by the corresponding SeOPD model and evaluate the PPL of the CoT using the Reference Model. As shown in Figure~\ref{fig:ppl_refkl}, the PPL of SeOPD-generated CoTs under the Reference Model generally increases throughout training across domains. Since the Reference Model remains fixed and PPL represents the extent to which the model understands the information within the text \citep{wan2024knowledge, marion2023less}, this trend indicates that SeOPD progressively generates CoT that is increasingly difficult for the initial model to predict, rather than remaining confined to its existing knowledge and expression patterns. This provides evidence that SeOPD progressively internalizes new information from CoT and subsequently leverages the acquired information to generate further content beyond the initial model's representational scope.

Finally, case studies are further provided to show how such new information emerges within the CoT (Appendix \ref{sec:case_study}).

% 自进化后的模型，用ref模型的cot，相比于ref，能力是否能提升？

% 注意力矩阵的rank是否有提升。（rank越大，容量越大）？ 或者有没有其它的指标衡量模型的知识量。

\subsection{Robustness to the Source of Privileged CoT (for Q4)}

We further investigate whether SeOPD depends on the source of privileged CoT. We fix the target model and the optimization procedure, while replacing the privileged CoT with CoTs generated by external LLMs with different scales and architectures. The results are reported in Figure~\ref{fig:cot_sources}.

CoTs from all sources effectively improve both Thinking and Non-Thinking performance, indicating that SeOPD is not tied to a specific CoT generator. Interestingly, the performance remains relatively stable across CoT sources with different model scales and capabilities, suggesting that the effectiveness of privileged CoT is not strongly dependent on the capability of the teacher model.

Notably, self-generated CoT is already highly competitive with external CoTs. This is partly because, unlike methods that use externally generated CoTs fixed throughout training, SeOPD regenerates Self-CoT using the evolving Thinking policy after each training step, providing continuously updated privileged information. In contrast to fixed offline CoT, this allows the privileged information to evolve together with the model rather than remaining unchanged throughout optimization. Overall, these results demonstrate that SeOPD can effectively leverage its own evolving reasoning process as internal PI without requiring a stronger external reasoning model.

\section{Conclusion}
We present SeOPD, a self-evolving framework that enables LLMs to improve their thinking and non-thinking policies using self-generated CoT as internal privileged information. By distilling reasoning knowledge from the model's own Thinking mode, SeOPD provides an alternative to externally provided answers, environment feedback, demonstrations, or handcrafted skills. Extensive experiments across science Q$\&$A, tool use, and coding demonstrate that SeOPD consistently improves both Thinking and Non-Thinking capabilities, while outperforming several methods relying on external privileged information. Our results highlight the potential of a model's own reasoning process as a valuable source of privileged information for self-evolution.

% \section{General formatting instructions}
% \label{gen_inst}

% The text must be confined within a rectangle 5.5~inches (33~picas) wide and
% 9~inches (54~picas) long. The left margin is 1.5~inch (9~picas).
% Use 10~point type with a vertical spacing of 11~points. Times New Roman is the
% preferred typeface throughout. Paragraphs are separated by 1/2~line space,
% with no indentation.

% Paper title is 17~point, in small caps and left-aligned.
% All pages should start at 1~inch (6~picas) from the top of the page.

% Authors' names are
% set in boldface, and each name is placed above its corresponding
% address. The lead author's name is to be listed first, and
% the co-authors' names are set to follow. Authors sharing the
% same address can be on the same line.

% Please pay special attention to the instructions in section \ref{others}
% regarding figures, tables, acknowledgments, and references.

% There will be a strict upper limit of \textbf{9 pages} for the main text of the initial submission, with unlimited additional pages for citations. This limit will be expanded to \textbf{10 pages} for rebuttal/camera ready.

\subsection*{Limitations}

First, our experiments primarily focus on relatively small and medium-sized dense LLMs. The scalability of SeOPD to substantially larger models and different architectures, such as Mixture-of-Experts (MoE) models, remains to be investigated.

Second, SeOPD currently relies on the model's existing Thinking capability to generate Self-CoT. Its effectiveness may therefore be constrained when the initial model has weak reasoning ability or lacks a meaningful distinction between Thinking and Non-Thinking modes. Investigating how to bootstrap SeOPD for models without strong initial reasoning capabilities is an interesting direction for future work.

\bibliography{iclr2027_conference}

\appendix
\section{Appendix}
\subsection{Experimental Settings}
\label{sec:expsetting}
\subsubsection{Environment}
Our training infrastructure consists of a single node with 8× NVIDIA H20 GPUs (96 GB HBM3 each), providing 768 GB of aggregate GPU memory. The software stack runs Python 3.13 on CUDA 12.4 with PyTorch. We build on the verl reinforcement learning framework \citep{sheng2024verl}, using PyTorch Fully Sharded Data Parallel (FSDP2) for parameter-efficient distributed training across all 8 GPUs. Rollout generation is handled by vLLM \citep{kwon2023vllm} with a tensor parallelism degree of 4, enabling high-throughput batched inference during on-policy data collection.

\subsubsection{hyperparams}
The hyperparameters used for SeOPD is summarized in Table \ref{tab:hyperparams}

\begin{table*}[ht]
\centering
\caption{Hyperparameters for SeOPD training.}
\label{tab:hyperparams}
\small
\begin{tabular}{l c c c}
\toprule
\textbf{Parameters} & \textbf{Science QA} & \textbf{ToolUse} & \textbf{LCB} \\
\midrule
\multicolumn{4}{l}{\textbf{Data}} \\
Max prompt length & 2048 & 2048 & 2048 \\
Max response length & 13312 & 13312 & 24576 \\
\midrule
\multicolumn{4}{l}{\textbf{Batching}} \\
Question batch size & 32 & 32 & 32 \\
Mini batch size & 32 & 32 & 32 \\
Number of rollouts & 8 & 8 & 8 \\
\midrule
\multicolumn{4}{l}{\textbf{Rollout}} \\
Inference engine & vLLM & vLLM & vLLM \\
Temperature & 1.0 & 1.0 & 1.0 \\
\midrule
\multicolumn{4}{l}{\textbf{Validation}} \\
Number of rollouts & 16 & 16 & 16 \\
Temperature & 0.6 & 0.6 & 0.6 \\
Top-$p$ & 0.95 & 0.95 & 0.95 \\
\midrule
\multicolumn{4}{l}{\textbf{Loss}} \\
$\alpha$ & 0.5 & 0.5 & 0.5 \\
Top-$K$ distillation & 100 & 100 & 100 \\
Teacher-EMA update rate & 0.0 & 0.0 & 0.0 \\
Rollout importance sampling clip & 2.0 & 2.0 & 2.0 \\
Online CoT KL coef & 0.01 & 0.01 & 0.01 \\
\midrule
\multicolumn{4}{l}{\textbf{Training}} \\
Optimizer & AdamW & AdamW & AdamW \\
Learning rate & $1 \times 10^{-5}$ (constant) & $1 \times 10^{-5}$ (constant) & $1 \times 10^{-5}$ (constant) \\
Warmup steps & 10 & 10 & 10 \\
Weight decay & 0.01 & 0.01 & 0.01 \\
Gradient Clip Norm & 1.0 & 1.0 & 1.0 \\
Hardware & $1 \times 8$ H20 & $1 \times 8$ H20 & $1 \times 8$ H20 \\
\bottomrule
\end{tabular}
\end{table*}

\subsubsection{Datasets}
We evaluate generalization on tasks on which the model has not been explicitly
fine-tuned, covering scientific reasoning, tool use, coding, mathematics,
and instruction following.
\paragraph{Science Q\&A.}
We use the undergraduate-level scientific reasoning subset (L3) of SciKnowEval \citep{sciknoweval2024},
covering Chemistry, Physics, Biology, and Materials Science.
L3 targets knowledge reasoning, including critical thinking, logical deduction,
numerical calculation, and scientific problem solving.
We report results on each domain as well as their mean.
\paragraph{Tool use.}
We evaluate tool calling on ToolAlpaca \citep{toolalpaca2023}.
Given a tool-API specification and a user request, the model must map the request
to the correct tool call.
This setting tests whether the model can follow structured interfaces that are
not seen during SeOPD training.
\paragraph{Coding.}
LiveCodeBench (LCB) \citep{livecodebench2024} provides contest-style programming problems ranging from
simple exercises to competition-level tasks.
We restrict evaluation to the most recent LCBv6 split, which contains 131
problems released between February and May 2025.
Following common practice on coding platforms such as LeetCode, we consider a
setting with both public and private unit tests:
public tests are used for evaluation during training, while private tests are
reserved for validation.
\paragraph{Hold-out tasks.}
To further probe generalization beyond the above domains, we evaluate two
hold-out benchmarks.
MATH-500 \citep{mathdataset2021} is a 500-problem subset of the competition-level MATH benchmark,
covering algebra, geometry, number theory, and related topics at high-school
contest difficulty.
IFEval \citep{ifeval2023} measures instruction following with about 500 prompts and 25 types of
automatically verifiable constraints, such as length limits, keyword inclusion,
and required output formats.
Neither benchmark is used for training, and both serve as hold-out tests of
mathematical reasoning and constraint satisfaction. We evaluate both the Thinking and Non-Thinking policies under the
same protocol.

\subsubsection{Evaluation protocol}
\label{sec:eval_protocol}
Unless otherwise specified, we report \textbf{mean@16}:
for each example we draw 16 independent completions and average
their binary correctness.

\subsubsection{The detail of the other OPSD methods}
\label{sec:other_method}

\textbf{SDFT}~\cite{shenfeld2026selfdistillation}
leverages few-shot prompts as privileged information (PI), which are
continuously optimized by an external LLM throughout training.
The model then performs on-policy distillation from these dynamically
updated prompts.

\textbf{SDPO}~\cite{hubotter2026reinforcement}
leverages two types of privileged information: answer information and
environment feedback. These sources of PI are converted into dense
token-level learning signals to guide on-policy policy optimization.

\textbf{OPCD}~\cite{ye2026onpolicycontext}
leverages external contextual information, such as optimized system
prompts, as privileged information and distills the additional
knowledge into the model parameters through on-policy training.

\textbf{U-OPSD}~\cite{li2026policy}
selects a response that agrees with the majority of sampled responses
through a consistency-based filtering mechanism and uses it as
privileged information for on-policy self-distillation. Since its
training code is not publicly available, we adopt SDPO as the
underlying training framework for our implementation.

\textbf{Skills}~\cite{miaoge2026competitiveprogrammingexpert}
injects programming knowledge from the \textit{Competitive Programming
Expert} skill into the prompt and adopts SDPO as the underlying
training framework for on-policy policy distillation.

\subsection{Dynamics of Fast-Thinking Length}
\label{app:fast_thinking_length}
\begin{figure*}[ht]
    \centering
    \includegraphics[width=0.92\linewidth]{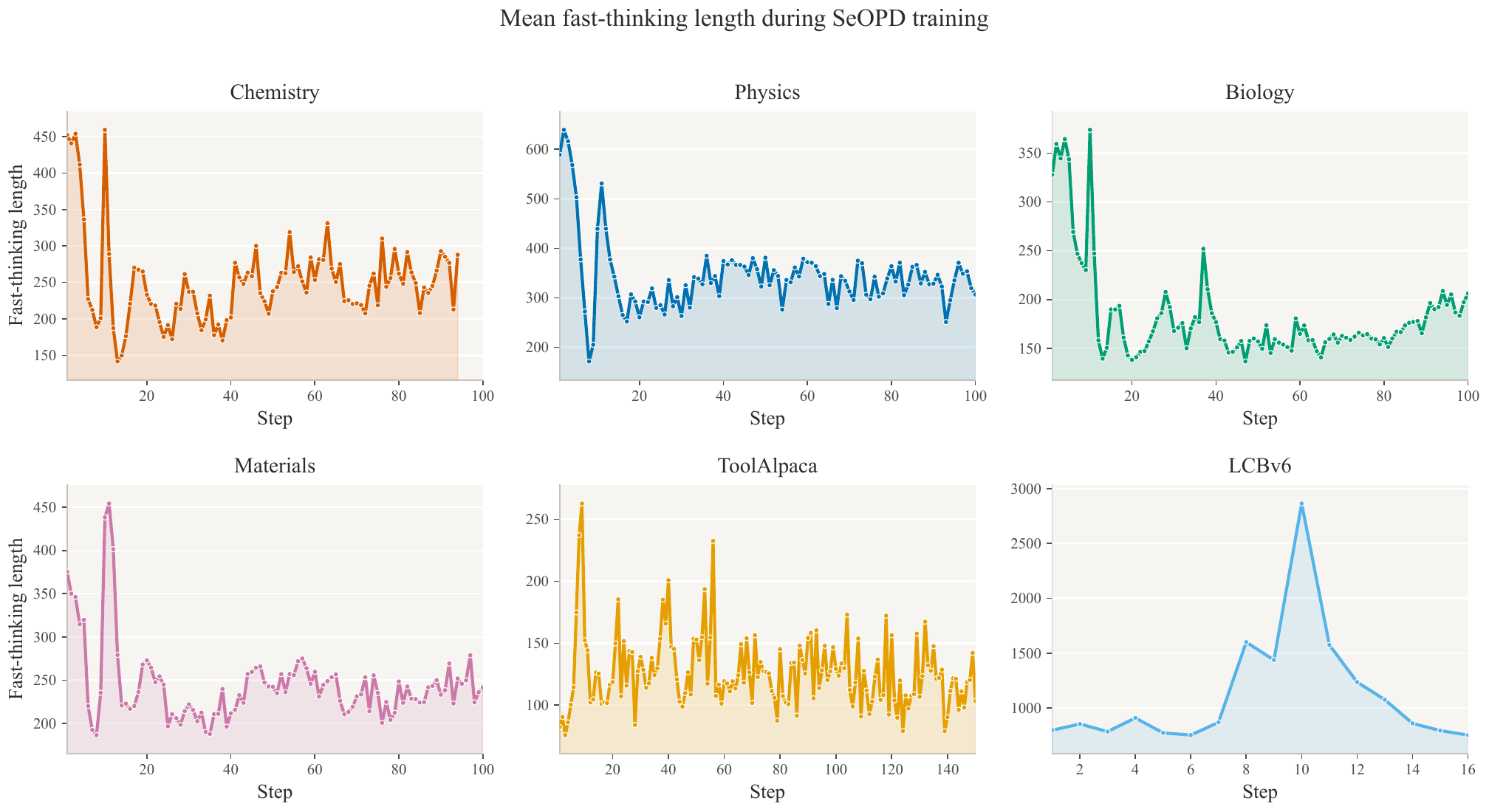}
    \caption{Mean response length of non-thinking during the first 40 SeOPD training steps. Each panel corresponds to one task. LCBv6 is shown up to step 19, which is the length of that run.}
    \label{fig:response_length}
\end{figure*}

SeOPD trains a Non-Thinking policy against online CoT
supervision. 
By tracking the mean response length of these Non-Thinking rollouts over
the training steps (Figure~\ref{fig:response_length}), we observe two
distinct trajectories that reveal how models internalize different types
of tasks: \emph{monotonic compression} and \emph{rise-and-fall structural mastery}.

\paragraph{Science Q$\&$A tasks: Monotonic compression via increased confidence.}
On knowledge-intensive and reasoning tasks (Chemistry, Physics, Biology, and Materials), the fast-thinking length exhibits a rapid, monotonic decrease after a short initial transient.
In the early stages of training, the student model lacks the internal
capability to solve complex problems directly. When forced to answer without CoT, it often struggles, generating verbose, exploratory, or repetitive text (e.g., restating the premise) to mask its uncertainty. As SeOPD proceeds, the student successfully absorbs the teacher's latent reasoning capabilities into its parameters. With this internalized knowledge, the model becomes confident. It sheds the redundant, hesitant verbosity and learns to output the concise final answer (e.g., a single option or value) directly and decisively.
This length reduction (37\%--55\%) serves as a behavioral signature of successful capability internalization.

\paragraph{Tool use and Coding: Rise-and-fall via structural mastery.}
In contrast, ToolAlpaca and LCBv6 exhibit an inverted-U trajectory: the length first spikes significantly before gradually decreasing. Unlike science QA, the target outputs for tool use and coding are inherently long and complex structural artifacts (e.g., complete JSON API calls or functional code blocks). We attribute this rise-and-fall dynamic to a two-phase learning process:
\begin{itemize}
    \item \textbf{Phase 1 (Rise - Mastering Structure):} The initial Non-Thinking policy often fails by generating truncated, syntactically invalid, or overly terse outputs (hence the low initial length). To achieve correctness, the model must first learn to construct the full required structure. During this phase, it rapidly expands its output length, sometimes over-generating redundant code paths or unnecessary parameters to ensure the structure is complete.
    \item \textbf{Phase 2 (Fall - Pruning Redundancy):} Once the policy masters the syntax and internalizes the core logic, it optimizes for efficiency. It begins to prune the suboptimal, redundant components generated in the first phase, distilling the output down to the most concise and effective API call or code snippet. The length thus falls back and stabilizes.
\end{itemize}
\paragraph{Summary.}
These contrasting dynamics highlight that ``internalization'' manifests
differently depending on the target output space. For tasks with concise answers, internalization replaces verbose hesitation with decisive brevity; for constructive tasks, it requires first expanding the output to master complex structures, followed by pruning to achieve efficiency.

\subsection{Case Study}
\label{sec:case_study}

\subsubsection{Case Study: How New Information Emerges within CoT}

To further illustrate how new information emerges within self-generated CoT, we present three representative case studies below.

In Table~\ref{tab:case_q1}, the pre-SeOPD CoT does not initially establish what constitutes a heavy atom and repeatedly considers different interpretations of the term. During reasoning, however, it eventually derives the key fact that heavy atoms refer to atoms other than hydrogen and correctly applies this knowledge to $\mathrm{C_4H_9NO_2}$, yielding seven heavy atoms. After SeOPD, the model directly recalls this knowledge and uses it in subsequent reasoning without re-deriving or reconsidering the definition. This case provides a concrete example of knowledge internalization: information first emerges within the CoT and is subsequently incorporated into the model, allowing later responses to directly leverage it during reasoning.

In Table~\ref{tab:case_q2}, the post-SeOPD CoT is reduced from 1,100 to 350 tokens while preserving the same tool-selection strategy. The model directly identifies the appropriate tool and maps the user's requirements to its arguments, without the lengthy verification and deliberation observed before SeOPD. This demonstrates that information acquired through CoT can also be internalized into more efficient procedural knowledge, enabling the model to reach the same action with substantially less explicit reasoning.

\setlength{\dblfloatsep}{8pt}
\begin{table*}[t]
\centering
\caption{Case Study Q1: Knowledge Internalization on Science Q\&A.}
\label{tab:case_q1}
\small
\renewcommand{\arraystretch}{1.05}
\setlength{\abovecaptionskip}{2pt}
\setlength{\belowcaptionskip}{2pt}
\begin{tabular}{p{0.96\linewidth}}
\hline
\rowcolor{gray!80}\begin{minipage}{\linewidth}\vspace{0.12em}\textcolor{white}{How many heavy atoms are there in the molecule with the IUPAC name `(3R)-3-aminobutanoate'? A: 7 B: 30 C: 31 D: 9}\vspace{0.12em}\end{minipage} \\
\hline
\rowcolor{blue!5} 
\textbf{CoT before SeOPD (Number of tokens = 4612):} \newline
... Each carbon atom can have atoms attached. For atom count, I need to count all heavy atoms, probably meaning atoms other than hydrogen, but in chemistry, "heavy atoms" often refer to carbon, nitrogen, oxygen, etc., excluding hydrogen. In the context of this question, "heavy atoms" likely means the carbon atoms, as is common in some contexts, but the options include numbers like 7, 30, 31, 9, so it might not be just carbon. Perhaps "heavy atoms" includes all atoms except hydrogen. But let's think about the molecule. In "3-aminobutanoate", the carbon chain has 4 carbons. The amino group is -NH2, so there is one nitrogen and two hydrogens. But the carbon from the amino group might be considered part of the molecule, but "3-aminobutanoate" likely means the amino is attached directly to the carbon, so the carbon is still part of the chain. Perhaps "3-aminobutanoate" is a way to name an amino acid, like 3-aminobutanoic acid. But "butanoate" might imply the acid form. I think "3-aminobutanoate" could be 3-aminobutanoic acid, which has four carbons in the chain. But "butanoate" might be ambiguous; perhaps it's a four-carbon chain with a carboxylic acid. But let's count the atoms. Perhaps "3-aminobutanoate" is the same as 3-aminobutanoic acid, which has molecular formula C4H9NO2. So, carbon atoms: 4 Hydrogen: 9 Nitrogen: 1 Oxygen: 2 But "heavy atoms" – if it's carbon, then 4, but 4 is not in the options. Options are 7, 30, 31, 9. ... Oh! 7 is option A. Perhaps "heavy atoms" means atoms other than hydrogen. ... So, I'll go with A. \\
\rowcolor{blue!5} \vspace{-0.8em} \noindent\rule{\linewidth}{0.4pt} \\
\rowcolor{blue!5} 
\textbf{CoT After SeOPD (Number of tokens = 2023):} \newline
... In chemistry, heavy atoms often refer to atoms with atomic number greater than hydrogen, so carbon, nitrogen, oxygen, etc. In this case, for C4H9NO2, the heavy atoms are C4, N1, O2, so 4+1+2=7. Yes, 7 heavy atoms. Option A is 7. ... So, answer is A. \\
\hline
\end{tabular}
\end{table*}

\begin{table*}[t]
\centering
\caption{Case Study Q2: Procedural Efficiency on ToolAlpaca.}
\label{tab:case_q2}
\small
\renewcommand{\arraystretch}{1.05}
\setlength{\abovecaptionskip}{2pt}
\setlength{\belowcaptionskip}{2pt}
\begin{tabular}{p{0.96\linewidth}}
\hline
\rowcolor{gray!80}\begin{minipage}{\linewidth}\vspace{0.12em}\linespread{1.08}\selectfont\textcolor{white}{\textbf{Context:} Your task is to answer the user's question using available tools. You have access to the following tools: generateAvatar: Create a unique avatar based on the user's preferences. ... \newline \textbf{Question:} My Discord server's current icon is getting old. I'd like to get a fresh look. Please generate a minimal styled server icon for my server with the ID "987654321" in JPG format.}\vspace{0.12em}\end{minipage} \\
\hline
\rowcolor{blue!5} 
\textbf{CoT before SeOPD (Number of tokens = 1100):} \newline
Okay, the user wants a fresh server icon for their Discord server with ID "987654321". They specifically asked for a minimal style in JPG format. \newline
First, I need to verify if the server ID is correct. Since the user provided it directly, I'll assume it's accurate unless instructed otherwise. The request is straightforward—they want a new icon generated using the available tools.\newline
The tool \texttt{generateServerIcon} is perfect for this. It takes the server ID, style preference, and format. The user specified "minimal" style and JPG format, so I'll map those directly to the parameters. \newline
I wonder if "minimal" is clear enough for the tool's interpretation. Maybe the user expects something like clean lines or a simple geometric shape. But since they didn't provide more details, I'll stick to their exact request. \newline
The output should be a URL, which they can use directly in Discord. No need for additional steps unless they ask for something else. \newline
This seems like a simple, one-step task. The user knows what they want, and the tools can deliver it. No need to overcomplicate things here. \\
\rowcolor{blue!5} \vspace{-0.8em} \noindent\rule{\linewidth}{0.4pt} \\
\rowcolor{blue!5} 
\textbf{CoT After SeOPD (Number of tokens = 350):} \newline
The user wants a new server icon for their Discord server. The tool \texttt{generateServerIcon} is perfect for this. I need to use the server ID they provided and specify the style and format. They mentioned "minimal styled," so I'll use "minimal" as the style. The format should be JPG. No other details are given, so I'll keep it simple. \\
\hline
\end{tabular}
\end{table*}

In Table~\ref{tab:case_q3}, the post-SeOPD CoT presents a more systematic characterization of the solution. It identifies the three possible choices for the product element, explains why commutativity reduces the six permutations to three conditions, and derives the resulting constant-time and constant-space complexity. These reasoning patterns become more explicit after SeOPD, indicating that the model can internalize useful structures from its reasoning process and subsequently apply them in a more systematic manner.

These cases provide qualitative evidence that self-generated CoT can serve as a source of information that is progressively incorporated into the model, rather than merely serving as a transient reasoning trace.

\makeatletter
\setlength{\@fptop}{0pt}
\setlength{\@dblfptop}{0pt}
\setlength{\@fpsep}{8pt}
\setlength{\@dblfpsep}{8pt}
\setlength{\@fpbot}{0pt plus 1fil}
\setlength{\@dblfpbot}{0pt plus 1fil}
\makeatother
\begin{table*}[p]
\centering
\caption{Case Study Q3: Structural Internalization on LCBv6.}
\label{tab:case_q3}
\small
\renewcommand{\arraystretch}{1.2}
\begin{tabular}{p{0.96\linewidth}}
\hline
\rowcolor{gray!80}\begin{minipage}{\linewidth}\vspace{0.12em}\textcolor{white}{\textbf{Question:} Given three integers $A_1,A_2,A_3$, determine whether they can be permuted so that $B_1 \times B_2 = B_3$.}\vspace{0.12em}\end{minipage} \\
\hline
\rowcolor{blue!5} 
\textbf{CoT before SeOPD (Number of tokens = 2162):} \newline
We are given three integers A1, A2, and A3, and we need to determine whether there exists a permutation (B1, B2, B3) such that B1 * B2 = B3.\newline\newline
Since there are three elements, there are 3! = 6 possible permutations. One straightforward approach is to explicitly enumerate all six permutations and check whether the first two elements multiply to the third. ...\newline\newline
Alternatively, because multiplication is commutative, we can reduce the six permutations to three distinct cases: A1 * A2 = A3, A1 * A3 = A2, or A2 * A3 = A1. ...\newline\newline
However, we should carefully verify that these three conditions indeed cover all possible permutations. For example, if the permutation is (A2, A1, A3), the condition becomes A2 * A1 = A3, which is equivalent to A1 * A2 = A3. Similarly, (A3, A1, A2) is equivalent to A1 * A3 = A2, and (A3, A2, A1) is equivalent to A2 * A3 = A1. ...\newline\newline
We also need to consider whether duplicate values cause any issue. The problem allows the input values to be equal, so a permutation should still use each occurrence exactly once. For example, for (2, 2, 4), the permutation (2, 2, 4) satisfies 2 * 2 = 4. Thus, duplicates do not change the validity of the three-condition formulation. ...\newline\newline
We could therefore either enumerate all permutations using itertools.permutations or directly check the three arithmetic conditions. Since the input contains only three integers, both approaches are easily fast enough. ...\newline\newline
Let me further verify the cases using the examples. For (3, 15, 5), we have 3 * 5 = 15, so the answer is Yes. For (5, 3, 2), none of the three products equals the remaining number, so the answer is No. ...\newline\newline
Therefore, the direct condition a*b == c or a*c == b or b*c == a is sufficient. ... \\
\rowcolor{blue!5} \vspace{-0.8em} \noindent\rule{\linewidth}{0.4pt} \\
\rowcolor{blue!5} 
\textbf{CoT After SeOPD (Number of tokens = 1100):} \newline
We need to determine whether the three given integers can be reordered so that the first two multiply to the third.\newline\newline
The key observation is that the permutation only determines which number serves as the product and which two numbers serve as its factors. Therefore, instead of explicitly constructing all six permutations, we can characterize the valid permutations by the three possible choices of the product element.\newline\newline
If A3 is the product, we need A1 * A2 = A3. If A2 is the product, we need A1 * A3 = A2. If A1 is the product, we need A2 * A3 = A1. Because multiplication is commutative, swapping the two factors does not create a new condition. Thus, these three checks completely characterize whether a valid permutation exists. ...\newline\newline
For example, with 3, 15, and 5, the third number can be viewed as the product because 3 * 5 = 15, so the corresponding permutation (3, 5, 15) satisfies the requirement. For 5, 3, and 2, none of the three possible product assignments is valid. ...\newline\newline
The input values are positive integers bounded by 100, so ordinary integer multiplication is sufficient. There are only three arithmetic comparisons, so the computation takes constant time and constant space. ...\newline\newline
Thus, the solution can directly evaluate:\newline\newline
a * b == c\newline
or\newline
a * c == b\newline
or\newline
b * c == a\newline\newline
If any of these conditions holds, we print "Yes"; otherwise, we print "No". The important part is that the permutation structure has been reduced to the underlying multiplicative relationship, so there is no need to explicitly enumerate or repeatedly verify the individual permutations. ... \\
\hline
\end{tabular}
\end{table*}

\end{document}